\documentclass[letterpaper, 10 pt, conference]{ieeeconf}  

\IEEEoverridecommandlockouts                              
\usepackage{graphicx}
\usepackage[hang,flushmargin]{footmisc}

\title{\LARGE \bf
Deep Tracking: Visual Tracking Using Deep Convolutional Networks
}

\author{Si Chen$^{1}$, Afshin Dehghan$^{2}$, and Meera Hahn$^{3}$
\noindent
\thanks{* All authors contributed equally to this work, authorship has been listed alphabetically.}
\thanks{$^{1}$
Department of Computer Science, The George Washington University, Washington, D.C. Contact: 
        {\tt\small sichen @ gwmail.gwu.edu}}%
\thanks{$^{2}$
Center for Research in Computer Vision, University of Central Florida, Orlando, Florida. Contact: 
        {\tt\small adehghan @ cs.ucf.edu }}%
\thanks{$^{3}$
Department of Computer Science, Emory University, Atlanta, Georgia. Contact: 
        {\tt\small mhahn7 @ emory.edu}}%
}

\begin{document}
\makeatletter
\let\latex@xfloat=\@xfloat
\def\@xfloat #1[#2]{%
  \latex@xfloat #1[#2]%
  \def\baselinestretch{1}
  \@normalsize\normalsize
  \normalsize
}
\makeatother
\maketitle
\thispagestyle{empty}
\pagestyle{empty}

\begin{abstract}

In this paper, we study a discriminatively trained deep convolutional network for the task of visual tracking. Our tracker utilizes both motion and appearance features  extracted from a pre-trained dual stream deep convolution network. We show that the features extracted from our dual-stream network can provide rich information about the target, leading to competitive performance against other state of the art tracking methods.
\end{abstract}

\section{INTRODUCTION}

Tracking is a crucial problem in computer vision and has been studied for decades. However, previous methods have not fully deciphered solutions to challenges in tracking such as illumination, occlusion, and scale. The recent application of deep learning methods have greatly improved performance in other computer vision problems such as object detection and action recognition. In this paper, we investigate the visual tracking problem from a deep learning approach. Moreover, we show that utilizing the deep features extracted from a pre-trained network produces a more effective and precise means for tracking.

Current state of the art trackers are able to address a few specific classical challenges each, such as scale or illumination. However, none are able universally handle the variety of issues that may occur in a given video. Handcrafted features such as color histogram, histogram of oriented gradients(HoG), and scale-invariant feature transform (SIFT)--the backbone of most previous trackers--are also prone to these problems. 

In this work we aim to investigate the use of recently developed deep features\cite{c1} in the context of tracking. Features extracted from a pre-trained deep network have shown to be reliable for many computer vision applications, such as object detection and action recognition. However, the usage of these features for visual tracking has not yet been explored. Here we propose a tracking pipeline which takes advantage of both appearance and motion features extracted from a pre-trained deep network. We show that the new features are capable of handling multiple of the common tracking challenges, such as illumination and occlusion and we show that it achieves better results compared to competitive approaches. 

\section{NETWORK STRUCTURES}
\subsection{Pipeline}
Our tracking algorithm starts by collecting positive and negative training samples from the video sequence. Positive samples contain the target whereas negative samples contain less than $20\%$ of the target. The tracker was given the bounding box of the target from the first video frame’s annotated ground truth, which included the center of the annotated bounding box as well as the width and height of the box. For the next $N$ frames, a simple tracker is used to track the location of the target in order to collect more positive training samples.

Using these N locations, we collected labeled positive and negative training samples through data augmentation. By permuting the bounding boxes through rotating and shifting the images several pixels each side, we collected 25 positive samples and 50 negative samples per frame. For the DT, N was set to four so that we could collect a total of 400 training samples.

Choosing the value of $N$ depends on the challenges faced in the first $N$ frames of a video sequence. For cases where the target did not experience a large deformation or illumination change in the first $N$ frames, lowering $N$ increased the efficiency of the tracker and produced better results. However, lowering $N$ results in a trade off yields a lower total number of training samples, decreasing accuracy in videos where the first frames do not provide a clear image of the target.

   \begin{figure}[thpb]
      \centering
      \includegraphics[scale=.225]{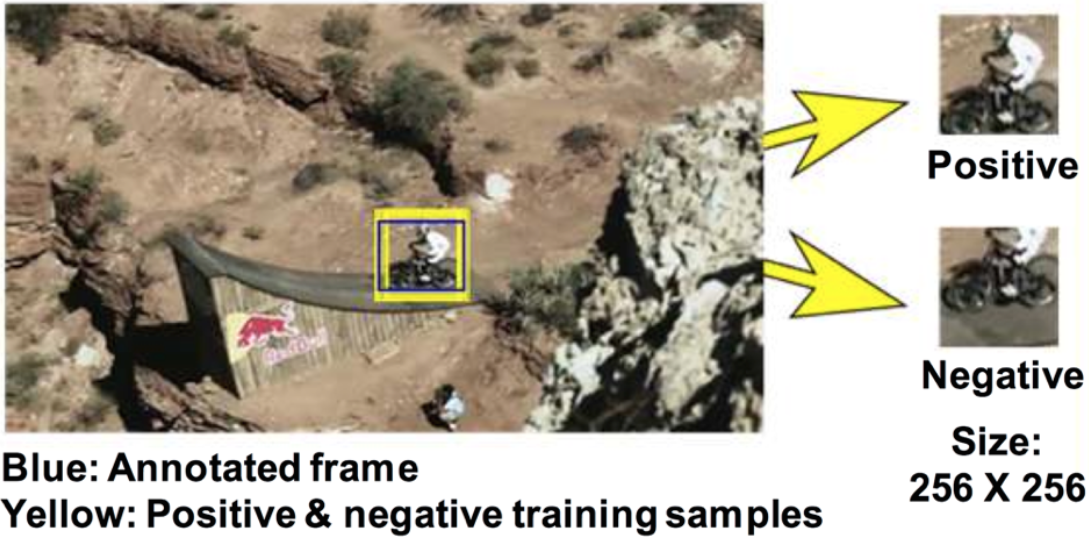}
      \caption{This figure shows an example of positive and negative samples generated given the first annotated bounding box of a video.}
      \label{figurelabel}
   \end{figure}
   
To collect testing samples for each frame, we took the location of the target in the previous frame and randomly chose test samples around that location using a Gaussian distribution since targets would not change location rapidly from frame to frame. The distribution addressed this by weighting the test samples near the last location more than test samples further away from the last location. 
\subsection{CNN with Pre-Training and SVM }
The initial tracking algorithm used a convolutional neural network (CNN) inspired by Krizhevsky’s network \cite{c5} to extract the features. The network has five convolutional layers and two fully connected layers with a max-pooling layer in between each convolution and a softmax regression loss.   

We obtained a model of the network that was pre-trained on the ImageNET  data set \cite{c6}. Our algorithm passed each of the labeled training images through the network and extracted the feature vector from the second fully connected layer. We passed these feature vectors to an SVM classifier for training. During testing, the same network is used to extract the features for every candidate window. Once the features are computed, the model parameters learned in training are passed to the SVM  for classification to get the confidence score for each sample. 

\subsection{Pre-Trained versus Fine-Tuned CNN}
The second CNN tested added an additional fully connected layer to the structure in II.B. This third fully connected layer had two outputs to match our two classes: background and foreground.In this model, the classifier and features were learned simultaneously to increase efficiency, a quality necessary in an online tracker. Furthermore, the low number of training samples for each video sequence prompted us to add fine-tuning for the three fully connected layers with our labeled training samples to the pre-trained model from II.B.

This approach was much simpler than re-training a CNN for each video sequence, given the low number of our training samples, and it made the tracking pipeline to be more efficient, a characteristic necessary in our goals to create an online tracking algorithm.

\section{Additional Pipeline Components}
\subsection{Optical Flow}
Motion information has shown to be a crucial component of tracking, especially when targets are occluded or have sudden changes in appearance. With a single stream network, our tracker is capable of extracting meaningful features, yet, due to common tracking problems such as occlusion, the  tracker would indicate that the foreground was moving to another spot much further away in the frame very suddenly. 

\begin{figure}[thpb]
      \centering
      \includegraphics[scale=.225]{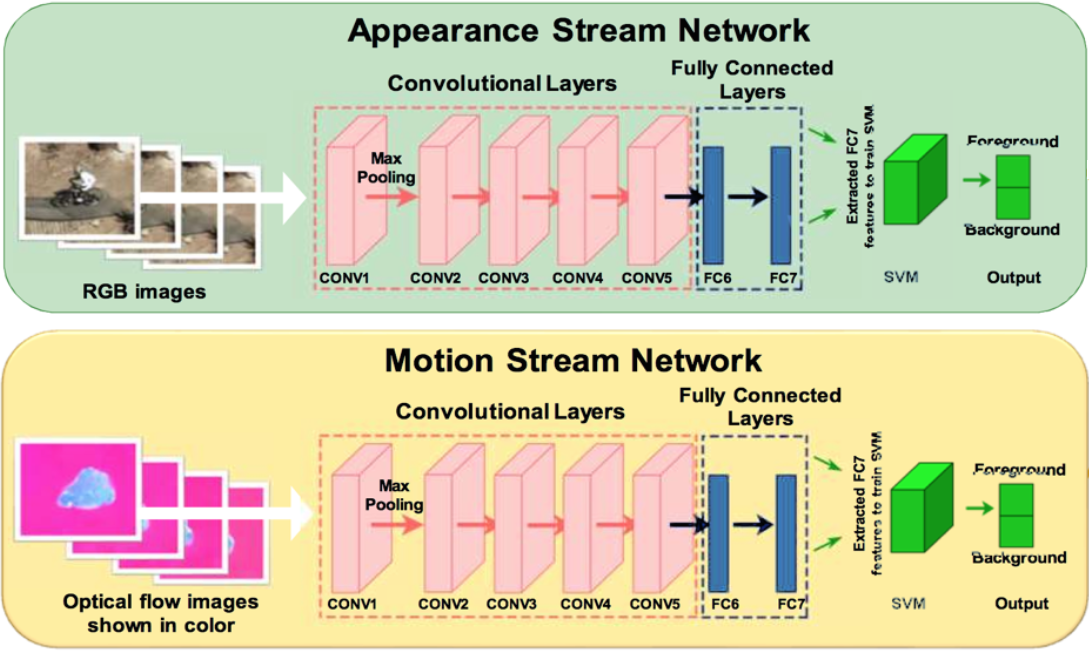}
      \caption{Fusion: Occurs after the SVM gives a confidence value and combines the outputs from both networks.}
      \label{fig:figureModel}
   \end{figure}

Our initial attempt to address this issue used a linear motion velocity model in the network described in II.B that did not prove to be effective. However, inspired by the work of Simonyan and Zisserman \cite{c7}, we use a double stream network that adds a motion stream network to our appearance stream network to incorporate temporal information. Our model is shown in Figure. \ref{fig:figureModel}.

For the motion stream network, we calculated optical flow between every adjacent two frames of the sequence and produced RGB optical flow images using Liu’s optical flow implementation \cite{c8}. We passed these RGB optical flow images into a network identical to the network described in II.C and trained a separate SVM classifier for the motion stream network. We ran both sets of test images through both networks and combined the scores output from the SVM using late fusion to take both the motion and  appearance of the target in the video into account when predicting the location of the object in the next frame. 

\subsection{Updating the Model}
The DT uses a two-stream network where the two streams are  updated  at different frequencies. The importance of updating the models can be seen in the issue of tracking objects whose appearance and direction of motion change. After qualitative observations that the direction and speed of the target changed more rapidly than its appearance, the motion stream network was updated more frequently than the appearance stream network. The motion stream network was updated every four frames whereas the appearance stream network was updated either every 50 frames or if the confidence values of the test images dropped below a defined threshold. This threshold was set to $0.85$. A value below $0.85$ indicates that a sudden occlusion or illumination change has occurred.

This method of updating the model allows the DT to handle challenges including over-fitting, change in motion, target shape, occlusion, and illumination.

\subsection{Handling Scale Change}
Handing  scale change is an important component of our tracking pipeline. If the target changes in size while the bounding boxes stay the same size, the resulting test images will contain a large amount of background information or only a portion of the target, leading to incorrectly labeled positive training samples.

From prior experimentation, we have found that the features our network learned were scale invariant to some extent and our network accurately handled the location of the target during the beginning of the scale change. Thus, updating the model based on a slight scale change was not necessary, and we could instead pass test samples that contain different scales to address scale change.  

Using the original methodology of collecting training and testing samples and then using the two-stream network described in III.A, the tracker chooses the location with the highest confidence score, assigning that as the location of the target.  At that chosen location, the image is scaled to 20 different sizes, which are passed as test images to the appearance network. The scale with the highest confidence score is chosen before the tracker updates the size of the bounding box. The main advantage of this approach is that we do not need to sample candidate from all  potential locations in different scales, thus reducing  computational complexity. 

\section{RESULTS}
\subsection{Data Sets}
The Deep Tracker was tested on the 315 video sequences from the Amsterdam Library of Ordinary Videos for tracking (ALOV++) \cite{c9} and the 29 video sequences from the Visual Tracker Benchmark \cite{c10}.

These data sets were chosen based on their videos’ diversity in circumstance--various combinations of classical computer vision problems such as occlusion, illumination, shape change, and low contrast were present in these videos.

Both data sets included the videos in single frames, the ground-truth annotations, and the results of the state of the art trackers that had been run on the data set. The ALOV++ data set had ground-truth annotations for every 5 frames and the Benchmark data set had ground-truth annotations for every frame. Both ground-truth annotations acknowledged scale change.

\subsection{F-Scores}
We evaluated our tracker against two state of the art tracking methods across two benchmark data sets. The scores are shown in the below tables. In the Visual Tracker Benchmark and the ALOV++ data set, our DT outperforms the state of the art tracker by at least 4.28\% and 2.77\% respectively.

    \section*{F SCORE COMPARISON}
    \section*{Visual Tracking Benchmark}
    \begin{center}
     \begin{tabular}{||c c c c||} 
     \hline
      & DT & Struck \cite{c12} & SCM \cite{c13} \\
     \hline
     Average & 65.92 & 55.19 & 61.64 \\ [1ex] 
     \hline
    \end{tabular}
    \end{center}
    \section*{ALOV++ Benchmark}
    \begin{center}
     \begin{tabular}{||c c c c||}
     \hline
       & DT & Struck & Difference \\
     \hline
     Average & 68.96 & 66.00 & 2.77 \\ [1ex] 
     \hline
     \end{tabular}
    \end{center}

\subsection{Success and Precision Plots}
Furthermore, we compared the DT to the top ten competitive methods using success and precision plots. These plots use receiver operating characteristic curves (ROC) to illustrate the inverse relationship between sensitivity and specificity. These curves were generated by plotting the comparison of true positives out of the number of total true positives and the fraction of false positives out of the total number of false positives. The success plot shows the overlap in the DT’s bounding boxes, the ground-truth, and the ratio of successful tracking, whereas the precision plot shows the error in the center location of the bounding box. The overall success and precision plots are shown below.
\\\\\\
\centerline{\includegraphics[scale=.22]{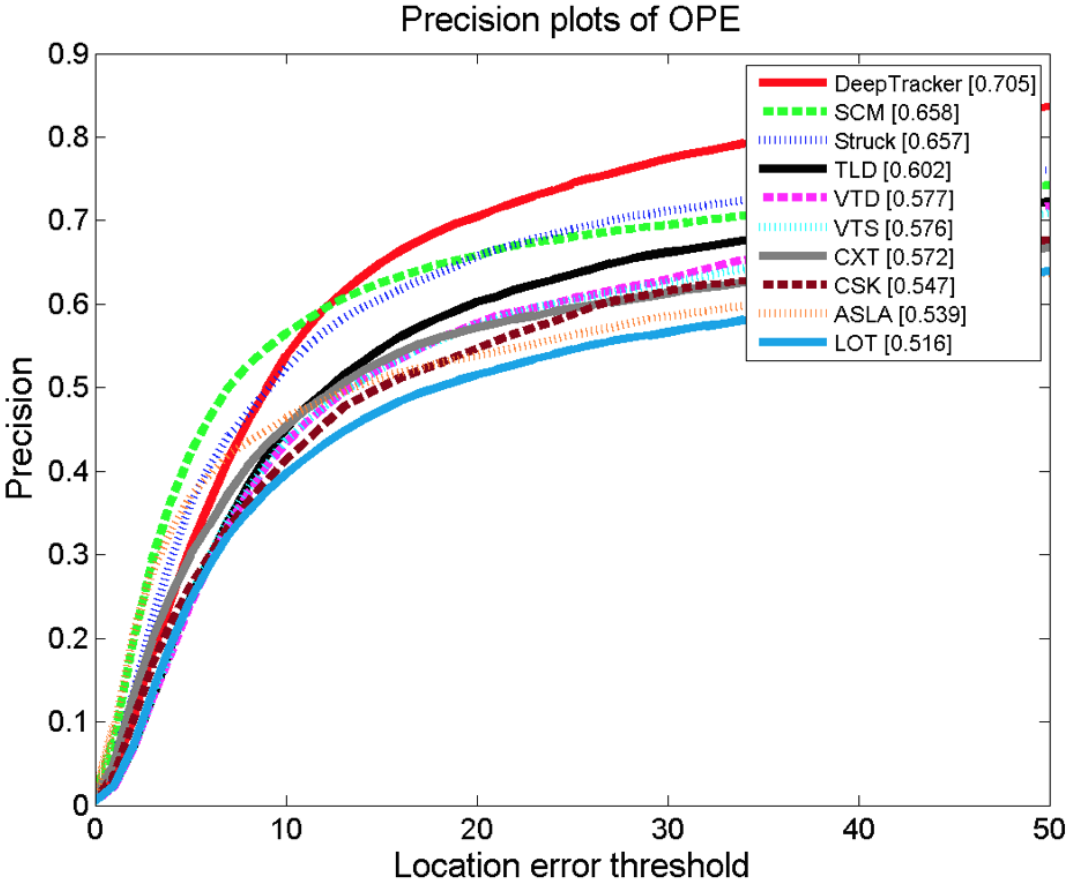}}
\vspace{1.5mm}

\centerline{\includegraphics[scale=.218]{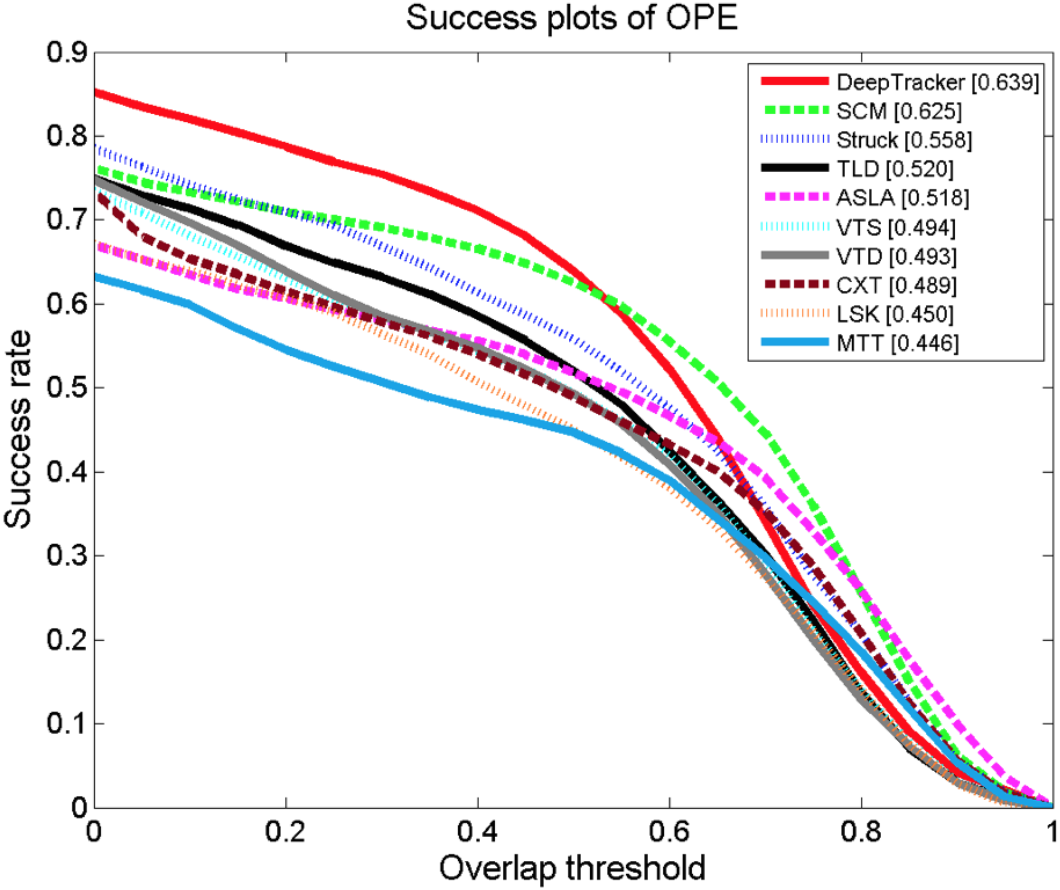}}
\vspace{1.5mm}

Success and precision plots were also calculated for specific classical computer vision problems. The DT outperformed state of the art tracking methods in several categories including occlusion, fast motion, background clutter, out of plane rotation, out of view, illumination variation, deformation, motion blur, and low resolution.
\section{CONCLUSION AND FUTURE WORK}
Our Deep Tracking algorithm takes advantage of deep features extracted from a pre-trained network, incorporating the information conveyed from both motion and appearance in a dual stream pipeline. Further expansions on this work will focus on training a CNN for the motion stream network using specifically RGB optical flow images.

Overall, our results show that the DT outperforms current state of the art tracking methods and encourage further work in the application of deep learning methods to tracking.\\\\

\addtolength{\textheight}{-12cm}   

\end{document}